\documentclass[letterpaper]{article} 
\usepackage[preprint]{aaai2027}  
\usepackage[hyphens]{url}  
\usepackage{graphicx} 
\usepackage{natbib}  
\usepackage{caption} 
\usepackage{booktabs}
\usepackage{array}
\usepackage{amsmath}
\usepackage{xcolor}
\usepackage{tikz}
\usetikzlibrary{arrows.meta,positioning}

\title{From Plausible Hierarchies to Useful Taxonomies: Evaluating Agentic Harnesses on Customer Feedback}

\author{Prabhath Chellingi, Raviraja G, Viraj Bagal \\ prabhath@enterpret.com, raviraja@enterpret.com, viraj@enterpret.com}
\affiliations{}

\begin{document}

\maketitle

\begin{abstract}
Taxonomies are the symbolic representations through which AI systems organize evidence, aggregate patterns, and answer questions over large document collections. Over customer feedback, the category tree decides how every record is counted and routed, which problems get seen, and which team owns them. Agentic harnesses now make it easy to generate a plausible-looking hierarchy, and such trees are checked today with \emph{generic}, individually scoped checks: each name fits its description, sits under the right parent, and stays distinct from its siblings. We ask a more operational question: when is a generated hierarchy actually useful as a production taxonomy? We build six taxonomies over two proprietary feedback corpora (1{,}940 and 5{,}000 records): for each corpus, a production reference and two repeated runs of the same harness under identical inputs. All six pass every generic naming and structure check, and a deeper product-coverage check even prefers the generated trees. Yet in every generated tree at least \textbf{97.7\%} of leaf names merely restate an ancestor's name (13.9\% and 2.9\% in the references), and in one, three of every four records fall under multiple top-level categories. We introduce two families of whole-tree metrics: \textbf{structural discriminators} test whether a tree's shape was learned from the data or imposed by its generator; \textbf{team partitionability} tests whether branches split feedback into groups teams can own. Trees the generic checks rate as equally correct differ by \textbf{27 percentage points} in cross-branch leakage, and only one of the two beats a random split. Surface plausibility is an insufficient measure of taxonomy quality: evaluation must measure the \emph{whole tree} as well as each node.
\end{abstract}


\section{Introduction}

A customer-feedback taxonomy has two jobs. It lets an organization ask ``which product area causes the most dissatisfaction'' and trust the count that comes back, and it lets a team scope its work down to the branch it owns. Neither job depends on any single node. Both depend on properties spread \emph{across} the tree: how volume distributes, how labels differ from their neighbors, how cleanly a branch cuts away from the rest.

Most taxonomy evaluation in use today does not look at the tree this way. It runs \emph{generic}, individually scoped checks, each inspecting a single node, edge, or sibling pair: does this name match this description, does this node belong under its parent, can these two siblings be told apart. These are legitimate questions, and an LLM judge answers them reliably. A few checks reach slightly further (sibling distinctness compares pairs, coverage checks compare nodes against documentation), but every verdict is still local, and the scores aggregate into a single pass-rate that cannot localize a systematic pattern. So a judge asked ``is this node's name coherent,'' one item at a time, never notices that every name in the tree is built from the same template, or that one whole level never branches. Those facts only exist at the level of the corpus.

We measured this gap on real artifacts. Over two proprietary customer-feedback corpora from two deployed commercial products (a small sample of 1{,}940 records and a large sample of 5{,}000) we assembled six taxonomies -- for each corpus, one production reference and two independent runs of the same off-the-shelf agentic construction harness under identical inputs and budgets -- and scored all six with the generic checks in common use. Every artifact passes every check. Description completeness sits at 1.00 across the board, name-description match at 0.97--1.00. Judged node by node, all six look comparably professional.

The whole-tree measurements disagree, and the failures repeat across runs and corpora. Every generated taxonomy restates a full ancestor name verbatim in \textbf{97.7--100\%} of its leaf names; the two references sit at 13.9\% and 2.9\%. In one generated taxonomy, every one of 158 parent nodes at one level has exactly one child -- a wrapper-node pattern that long-tailed feedback data simply does not produce. Routing the actual feedback records through these trees, between a third and three-quarters of each generated taxonomy's records land in more than one top-level product area, against a fifth for the references, and one generated taxonomy's branch boundaries score \emph{worse than a random partition} at keeping related issues together. The generic checks detect none of this. They score all six artifacts as comparably professional on naming and structure.

We make three contributions:

\begin{enumerate}
\item We show, with concrete numbers and taxonomy snippets, that a widely-used style of taxonomy evaluation (generic LLM judging of names, descriptions, edges, and sibling pairs) systematically fails to detect templated-label collapse, generator-imposed branching caps, and cross-team fragmentation -- the failure modes every agentic run in our study exhibits -- because none of these is a property of any single node.
\item We introduce two deterministic, judge-free metric families that are computable from the taxonomy artifact (and, for one family, its record classifications) alone: \textbf{structural discriminators}, which detect when structure was imposed by a generator rather than learned from data, and \textbf{team partitionability}, which measures whether a taxonomy's branches partition the record stream into cleanly separable groups, a structural prerequisite for assigning branches to distinct product teams and a downstream question existing evaluations do not ask.
\item We report both metric families over six independently constructed taxonomies on two corpora, show how to read each number against a concrete taxonomy snippet, and show that the new metrics separate artifacts the generic checks cannot: C-1 and C-2, two runs of the same harness that those checks rate as comparably correct (0.92--1.00 hierarchy correctness), differ by 27 percentage points in branch leakage and flip sign in co-occurrence modularity.
\end{enumerate}

\section{Related Work}

\subsection{Statistical and Corpus-Based Taxonomy Construction}

Traditional taxonomy-construction research has used term extraction, semantic similarity, clustering, and hierarchical expansion. TaxoGen recursively clusters terms using adaptive embeddings to construct topical taxonomies from corpora \citep{zhang2018taxogenunsupervisedtopictaxonomy}. HiExpan expands a seed taxonomy using corpus evidence and task guidance \citep{Shen2018HiExpanTT}. TaxoExpan expands an existing taxonomy with new concepts using a position-enhanced graph neural network trained on self-supervised anchor relations \citep{Shen2020TaxoExpanST}. Topic-modeling methods such as latent Dirichlet allocation \citep{Blei2009LatentDA} and embedding-based approaches such as BERTopic \citep{Grootendorst2022BERTopicNT} are commonly used to organize unstructured records. None of these methods is evaluated on whether its output partitions the underlying records into groups separable enough to divide among teams, the question our second metric family targets.

\subsection{Taxonomies for Customer Feedback}

Customer reviews, surveys, app-store reviews, and support interactions have been organized using topic and aspect taxonomies. Amazon's InsightNet constructs a multi-level taxonomy from customer reviews and associates granular topics with polarity and supporting verbatim \citep{Mukku2024InsightNetS}. Related work uses hierarchical classification to identify actionable topics in customer questions \citep{Rana2023WeaklySH}, and FeedbackMap organizes open-ended survey responses for analysis \citep{Beeferman2023FeedbackMapAT}. FeClustRE organizes extracted app features into hierarchically clustered and semantically labeled structures \citep{Tiessler2025FeClustREHC}. A systematic review of user-feedback classifications found a large, heterogeneous label space organized around sentiment, intention, user experience, and topic \citep{Santos2019ATF} -- itself evidence that no single taxonomy follows automatically from a collection of feedback. This line of work evaluates classification accuracy or topic coherence, not whether the resulting structure is fit for organizational use.

\subsection{LLM and Agentic Taxonomy Construction}

Pretrained language models have been used to construct taxonomic trees by predicting parenthood relations and reconciling them into a consistent hierarchy \citep{Chen2020ConstructingTF}, and instruction-tuned models have been applied to entity-set expansion, taxonomy expansion, and seed-guided taxonomy construction within a unified framework \citep{Shen2024AUT}. In low-resource settings, self-supervised prompting with reinforcement-learning fine-tuning expands taxonomies from a handful of labeled examples \citep{Mishra2024FLAMESL}. Agentic systems go further: gathering information, invoking tools, evaluating intermediate artifacts, and revising outputs across a multi-step trajectory. Recent evidence shows agentic benchmark performance is itself noisy across repeated runs of the identical task \citep{bjarnason2026randomnessagenticevals}; our results extend this to taxonomy construction, and show that repeated-run variance concentrates in the corpus-level properties generic evaluation cannot see.

\subsection{Taxonomy Quality Evaluation}

Prior work has proposed quality attributes for taxonomies -- comprehensiveness, robustness, explanatory power, extensibility, mutual exclusiveness -- historically assessed by expert judgment; a compendium of such attributes, itself contributing explicit measurements for them, shows that evaluation is inherently tied to intended use \citep{Unterkalmsteiner2022ACA}. Formal ontology evaluation offers complementary machinery: OntoClean audits subsumption relations against metaproperties \citep{Guarino2002EvaluatingOD}, an influential taxonomy-development method defines taxonomies over mutually exclusive and collectively exhaustive dimensions \citep{Nickerson2013AMF}, and surveys catalogue structural and application-based measures \citep{Brank2005ASO}. TExEval scores extracted taxonomy edges against gold-standard hierarchies \citep{Bordea2015SemEval2015T1,Bordea2016SemEval2016T1}, TnT-LLM evaluates LLM-generated label taxonomies with LLM judges at scale \citep{Wan2024TnTLLMTM}, and hierarchical-classification research contributes hierarchy-aware measures \citep{Kosmopoulos_2014}. \textbf{With one partial exception (mutual exclusiveness, which we revisit in the Discussion), every one of these instruments is scored by inspecting nodes, edges, or node pairs.} None asks whether a node's label is informationally distinct from its ancestors' labels in aggregate across the tree, whether a level's branching factor reflects the data's distribution or a generation artifact, or whether the tree's branches partition the underlying record stream into cleanly separable groups. Computational linguistics distinguishes word types from word tokens \citep{Jurafsky2000SpeechAL}, and their ratio is a classical measure of a text's vocabulary diversity; network science has long used modularity to measure whether a graph partition is meaningfully self-contained \citep{Newman_2006}. Neither tool has, to our knowledge, been applied to taxonomy artifacts. Goodhart's Law names the eventual risk: a measure optimized in isolation stops indicating what it proxied \citep{manheim2019categorizingvariantsgoodhartslaw}. The nearer problem we document is that node-level checks fail even to \emph{detect} these structures arising by default, before any optimization pressure exists.

\section{Experimental Setup}
\label{sec:setup}

Every number in this paper comes from the protocol described in this section -- the same six artifacts, the same two record sets, one of four scoring instruments per metric. That covers Table~\ref{tab:existing}'s generic check scores, Table~\ref{tab:structural}'s structural discriminators, Table~\ref{tab:partition}'s partitionability metrics, and every snippet in the figures.

\subsection{Corpora and Artifact Construction}
\label{sec:corpus}

We evaluate on customer-feedback corpora from two deployed commercial products, referred to throughout as the \emph{small sample} (1{,}940 records) and the \emph{large sample} (5{,}000 records); each corpus is accompanied by the product's own documentation. Both are proprietary, and we do not characterize the products or data pipelines further. Over each corpus we assembled three taxonomies, six artifacts in total:

\textbf{Live} (one per corpus) is that product's existing production taxonomy, built by the data owner's in-house pipeline (itself agentic, not hand-curated) around deliberate structural checks: recurring-pattern checks, documentation cross-referencing, sibling-overlap resolution. The Live taxonomies are reference points, not controlled arms: they ran on different infrastructure without a matched budget, and because their pipeline was engineered around checks similar in spirit to our metrics, their favorable profiles are partly by construction. The confound-free evidence in this paper is therefore the within-harness contrast among the generated runs; the references show what the metrics report on trees shaped by deliberate structural checks.

\textbf{C-1, C-2} (small sample) and \textbf{D-1, D-2} (large sample) are independent runs of the same off-the-shelf agentic coding harness (Codex \citep{openaiCodexChatGPT}, running GPT-5.6-Sol at \texttt{max} reasoning effort) under one fixed protocol: a fresh, isolated workspace per run; an identical task specification (reproduced in full in the Supplementary Material) describing the five-level schema, connectivity rules, a no-duplication rule. No run had access to the evaluator or to how it would be scored. Per the run manifests, the four runs consumed 5.8M--12.2M input tokens (96.9--98.1\% served from cache; fresh input 159K--296K) and 73K--95K output tokens each, of which 30.9--37.4\% were reasoning tokens.

Each artifact classifies its corpus's records via \texttt{predictions.jsonl}, with every record assigned one or more complete L1-to-Subtheme paths in that artifact's own tree. The records are identical across artifacts on the same corpus; the classifications are necessarily artifact-specific, since a path can only reference its own tree's nodes -- a dependency we return to in Section~\ref{sec:newmetrics} and Limitations. Substantive node counts (content-bearing nodes, excluding the system-managed generic/miscellaneous catch-alls that absorb unmatched records) range from 1{,}421 to 1{,}509 on the small sample and 4{,}700 to 4{,}829 on the large; the tables report rate-based statistics alongside raw counts so artifact size does not confound the comparison.

\subsection{Evaluation Protocol}
\label{sec:protocol}

Four scoring instruments produce the numbers in this paper; the two metric families we introduce use no language model:

\textbf{Generic checks (Table~\ref{tab:existing}).} Taxonomy quality is commonly assessed with generic checks scored by an LLM judge at scale \citep{Wan2024TnTLLMTM}, echoing the node-, edge-, and pair-level attributes prescribed in the taxonomy-evaluation literature (\S2.4). The judge reads one node, one edge, or one sibling pair at a time and returns a verdict; the reported score is the fraction of items passing. We evaluate four widely used checks of this kind (judge: gpt-5.4-mini), covering naming and structure, the axes Sections~\ref{sec:existing} and~\ref{sec:newmetrics} turn on:
\begin{itemize}
\item \emph{Name-description match}: the fraction of nodes whose name the judge finds consistent with that node's own description.
\item \emph{Description completeness}: the fraction of nodes carrying a non-empty, judged-adequate description.
\item \emph{Hierarchy correctness}: the fraction of parent-child edges the judge does \emph{not} flag as a wrong-parent relation.
\item \emph{Sibling distinctness}: the fraction of sibling pairs (nodes sharing an immediate parent) the judge does \emph{not} flag as overlapping or duplicative.
\end{itemize}
Each is judged one node, edge, or pair at a time -- never against an ancestor, never against the tree as a whole -- and this is the limitation Section~\ref{sec:existing} exploits.

\textbf{Product-term coverage (Table~\ref{tab:existing}, bottom row).} A deeper check of the kind commonly used to validate that a taxonomy covers the product it describes: canonical product terms are extracted from the product's documentation, and a stronger judge (gpt-5.5) decides per term whether some node in the taxonomy represents it; the score is the fraction of judged terms covered. It measures whether the taxonomy covers the product's documented surface, not whether the tree is internally sound. We report it because it is the one check on which the generated artifacts \emph{beat} the references (Section~\ref{sec:existing}).

\textbf{Structural discriminators (Table~\ref{tab:structural}).} Computed by a standalone script operating on \texttt{taxonomy.json} alone -- no predictions, no LLM call. Ancestor containment, token redundancy, and type-token ratio are string and set operations over node names (\S\ref{sec:newmetrics}); fan-out shape and cross-branch reuse are computed directly from the edge list.

\textbf{Team partitionability (Table~\ref{tab:partition}).} Computed by a second standalone script operating on \texttt{taxonomy.json} and \texttt{predictions.jsonl} together: every record's predicted paths are routed through each candidate tree to build the branch-assignment and leaf co-occurrence structures the leakage, purity, and modularity formulas (\S\ref{sec:newmetrics}) are computed over.

\section{Existing Metrics: What They Test and Where They Break}
\label{sec:existing}

\subsection{Generic Checks in Common Use}

Table~\ref{tab:existing} reports the four generic naming-and-structure checks, plus the deep-judged product-term coverage check, over the six artifacts of Section~\ref{sec:corpus}.

\begin{table}[t]
\centering
\small
\setlength{\tabcolsep}{1.5pt}
\begin{tabular}{@{}lcccccc@{}}
\toprule
& \multicolumn{3}{c}{Small sample} & \multicolumn{3}{c}{Large sample} \\
\cmidrule(lr){2-4} \cmidrule(lr){5-7}
\textbf{Metric}\footnotemark & Live & C-1 & C-2 & Live & D-1 & D-2 \\
\midrule
Name-desc.\ match $\uparrow$ & 0.98 & 0.97 & 0.98 & 0.98 & 1.00 & 1.00 \\
Desc.\ completeness $\uparrow$ & 1.00 & 1.00 & 1.00 & 1.00 & 1.00 & 1.00 \\
Hierarchy correctness $\uparrow$ & 0.94 & 1.00 & 0.92 & 0.93 & 0.94 & 0.99 \\
Sibling distinctness $\uparrow$ & 0.94 & 0.88 & 0.94 & 0.95 & 0.91 & 0.99 \\
\midrule
Product-term coverage $\uparrow$ & 0.72 & \textbf{0.74} & \textbf{0.86} & 0.80 & \textbf{0.91} & \textbf{0.90} \\
\bottomrule
\end{tabular}
\caption{The generic checks (top four rows) and the deep-judged product-term coverage check (bottom row), six artifacts across two corpora. Every artifact passes every check, with name-description match at 0.97--1.00 everywhere; the checks read as ``all six are professional-grade.'' On product-term coverage, \textbf{every generated artifact outscores its reference} (bold), on both corpora.}
\label{tab:existing}
\end{table}
\footnotetext{$\uparrow$ higher is better; every metric in this table is higher-is-better.}

On its face, Table~\ref{tab:existing} is reassuring: every artifact clears every check, on both corpora, and the generated runs frequently outscore their references (D-1 and D-2 post perfect 1.00 naming scores against their reference's 0.98). Product-term coverage goes further. Every generated artifact outscores its reference (0.74--0.86 against 0.72 on the small sample; 0.90--0.91 against 0.80 on the large), so by this check the generated trees look \emph{better} grounded in the product than the production taxonomies. Anyone reading this table alone would conclude that the generated trees are interchangeable with the production references, or better.

\subsection{What the Table Does Not Show}

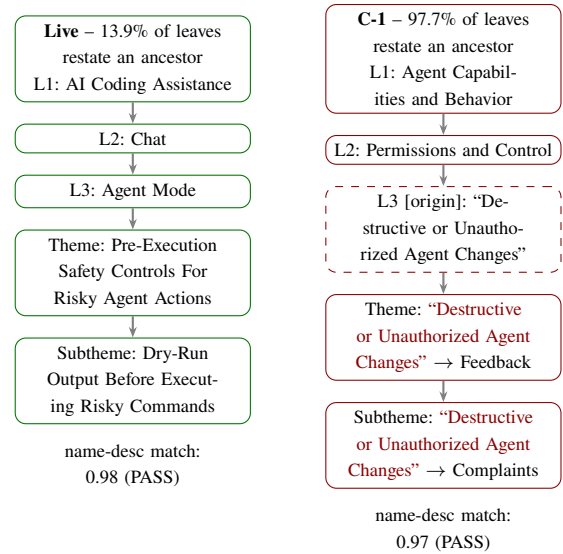
\begin{figure}[t]
\centering
\begin{tikzpicture}[
  every node/.style={font=\scriptsize},
  box/.style={draw, rounded corners, text width=2.9cm, align=center, inner sep=3pt},
  good/.style={box, draw=green!45!black},
  bad/.style={box, draw=red!55!black},
  origin/.style={bad, dashed},
  arr/.style={-{Stealth[length=4pt]}, thick, gray}
]
\node[good] (f1) {\textbf{Live} -- 13.9\% of leaves restate an ancestor\\L1: AI Coding Assistance};
\node[good, below=8pt of f1] (f2) {L2: Chat};
\node[good, below=8pt of f2] (f3) {L3: Agent Mode};
\node[good, below=8pt of f3] (f4) {Theme: Pre-Execution Safety Controls For Risky Agent Actions};
\node[good, below=8pt of f4] (f5) {Subtheme: Dry-Run Output Before Executing Risky Commands};
\draw[arr] (f1)--(f2); \draw[arr] (f2)--(f3); \draw[arr] (f3)--(f4); \draw[arr] (f4)--(f5);
\node[below=4pt of f5, font=\scriptsize, text width=2.9cm, align=center] {name-desc match: 0.98 (PASS)};

\node[bad, right=28pt of f1] (c1) {\textbf{C-1} -- 97.7\% of leaves restate an ancestor\\L1: Agent Capabilities and Behavior};
\node[bad, below=8pt of c1] (c2) {L2: Permissions and Control};
\node[origin, below=8pt of c2] (c3) {L3 [origin]: ``Destructive or Unauthorized Agent Changes''};
\node[bad, below=8pt of c3] (c4) {Theme: \textcolor{red!55!black}{``Destructive or Unauthorized Agent Changes''} $\to$ Feedback};
\node[bad, below=8pt of c4] (c5) {Subtheme: \textcolor{red!55!black}{``Destructive or Unauthorized Agent Changes''} $\to$ Complaints};
\draw[arr] (c1)--(c2); \draw[arr] (c2)--(c3); \draw[arr] (c3)--(c4); \draw[arr] (c4)--(c5);
\node[below=4pt of c5, font=\scriptsize, text width=3.0cm, align=center] {name-desc match: 0.97 (PASS)};
\end{tikzpicture}
\caption{One leaf path each from Live and C-1, both real, verbatim node names; column headers give each artifact's measured ancestor-containment rate. The phrase in quotation marks (set in red) is \textbf{copied character-for-character} from C-1's L3 node into every descendant below it: Theme and Subtheme are each almost entirely copied text with a single new word appended. Name-description match passes both chains at 0.97--0.98 (Table~\ref{tab:existing}), because it only ever compares a node to its own description, never to its ancestors.}
\label{fig:existing-fail}
\end{figure}

Figure~\ref{fig:existing-fail} shows one leaf path from Live and from C-1 side by side, each column headed by that artifact's measured ancestor-containment rate (the fraction of leaves whose name restates a full ancestor name; defined formally in Section~\ref{sec:newmetrics}), with the name-description score for that leaf annotated underneath. C-1 does pass name-description match (0.97, matching Table~\ref{tab:existing}), so the check is not quietly catching the problem some other way. C-1's L3 node (dashed border) is where the repeated phrase originates; every box below it re-pastes that same quoted L3 name, verbatim, into its own name before appending one new word. The repetition is not a presentation choice in the figure -- the names are stored this way in \texttt{taxonomy.json}.

A judge scoring \emph{one node at a time} -- ``does this name match this description'' -- has no way to see this. Both leaves have coherent names that match their own descriptions perfectly well in isolation, and the generic checks score both artifacts as comparably professional. The difference only appears when one looks at \emph{all} of an artifact's leaves at once and asks what fraction of them are built this way. C-1 restates a matched ancestor's name in \textbf{97.7\% of its 867 leaves} (Table~\ref{tab:structural}); C-2, not pictured, reaches \textbf{100\% of its 599 leaves}, and the same pattern recurs independently on the large sample (97.7\% and 100\%). Nothing in Table~\ref{tab:existing} hints at any of this; the failure sits underneath clean 0.97--1.00 naming scores. The two references, for comparison, restate an ancestor's name in only \textbf{13.9\%} and \textbf{2.9\%} of their leaves.

The same blindness recurs for branching structure. C-2's L2-to-L3 fan-out is the most extreme version of a generator-imposed branching cap we observed: all 158 of its L2 parents have exactly one L3 child, a pure chain of wrapper nodes with no variation at all, while its widest node at any \emph{other} level pair still reaches 26 children, against 31 for Live. Hierarchy correctness (Table~\ref{tab:existing}) scores C-2 at 0.92, because each of its edges, judged individually, looks like a defensible parent-child relation. What that metric cannot see is that the \emph{overall shape} of C-2's branching was clamped by the generation process rather than shaped by where the feedback actually concentrates. Figure~\ref{fig:fanout} draws the contrast.

What the two failures have in common is that they are true of the whole tree, not of any node the generic checks inspect -- and both artifacts still score highly on every generic check in Table~\ref{tab:existing}, while outscoring the reference on product-term coverage.

\section{Two New Metric Families}
\label{sec:newmetrics}

We propose two metric families computed directly from the taxonomy artifact (and, for the second family, the record classifications) with no LLM judge and no gold taxonomy required, following the evaluation protocol of Section~\ref{sec:protocol} over the same six artifacts and record sets as Table~\ref{tab:existing}.

\subsection{Structural Discriminators}

These metrics ask whether the tree's structure was learned from the data or imposed by the generator.

\textbf{Label information gain.} For a non-root node $n$ with ancestor set $A(n)$ (system-managed catch-all nodes excluded) and $\mathrm{tok}(x)$ the content words in node $x$'s name (the words remaining after stopword removal), \emph{ancestor containment} flags $n$ when a full multi-word ancestor name is a substring of $n$'s name; \emph{token redundancy} is the continuous version:
\[
\mathrm{redund}(n) = \frac{|\mathrm{tok}(n) \cap \mathrm{tok}(A(n))|}{|\mathrm{tok}(n)|}
\]
\[
\mathrm{gain}(n) = 1 - \mathrm{redund}(n)
\]
High containment with low mean $\mathrm{gain}(n)$ means child labels are restating rather than specializing their parents.

\textbf{Lexical diversity.} Per level $L$, the type-token ratio -- distinct words (types) divided by total words (tokens) \citep{Jurafsky2000SpeechAL} -- over node names at that level:
\[
\mathrm{TTR}_L = \frac{|\{\text{unique tokens in names at } L\}|}{|\{\text{total tokens in names at } L\}|}
\]
A generator recycling a small set of template phrases collapses this several-fold relative to organically written labels. TTR generally declines as a corpus grows, and Live has the most leaves yet the highest TTR in our study, so the reported gaps are, if anything, conservative.

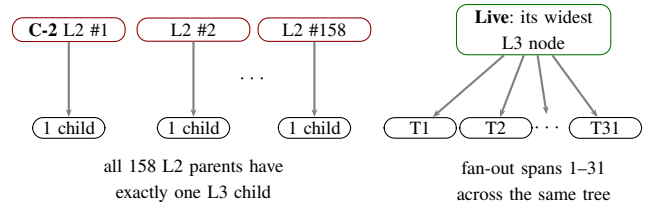
\begin{figure}[t]
\centering
\begin{tikzpicture}[
  every node/.style={font=\scriptsize},
  pbox/.style={draw, rounded corners, text width=1.35cm, align=center, inner sep=2pt},
  cbox/.style={draw, rounded corners, text width=0.85cm, align=center, inner sep=1.5pt},
  arr/.style={-{Stealth[length=3pt]}, thick, gray}
]
\node[pbox, draw=red!55!black] (p1) at (-1.3,0) {\textbf{C-2} L2 \#1};
\node[pbox, draw=red!55!black] (p2) at (0.35,0) {L2 \#2};
\node[pbox, draw=red!55!black] (p3) at (1.95,0) {L2 \#158};
\node[cbox] (a1) at (-1.3,-1.3) {1 child};
\node[cbox] (a2) at (0.35,-1.3) {1 child};
\node[cbox] (a3) at (1.95,-1.3) {1 child};
\node[font=\scriptsize] at (1.15,-0.65) {$\cdots$};
\draw[arr] (p1)--(a1); \draw[arr] (p2)--(a2); \draw[arr] (p3)--(a3);
\node[font=\scriptsize, text width=2.9cm, align=center] at (0.35,-2.0) {all 158 L2 parents have exactly one L3 child};

\node[pbox, draw=green!45!black, text width=1.9cm] (q1) at (4.85,0) {\textbf{Live}: its widest L3 node};
\node[cbox] (b1) at (3.35,-1.3) {T1};
\node[cbox] (b2) at (4.35,-1.3) {T2};
\node[font=\scriptsize] (bdots) at (5.05,-1.3) {$\cdots$};
\node[cbox] (b3) at (5.8,-1.3) {T31};
\draw[arr] (q1)--(b1); \draw[arr] (q1)--(b2); \draw[arr] (q1)--(bdots); \draw[arr] (q1)--(b3);
\node[font=\scriptsize, text width=2.6cm, align=center] at (4.85,-2.0) {fan-out spans 1--31 across the same tree};
\end{tikzpicture}
\caption{Fan-out shape (child counts real; node identities schematic). C-2's L2$\to$L3 level pair collapses to a pure wrapper chain: \textbf{all 158 parents have exactly one child}. Live's fan-out spans 1--31, the shape long-tailed feedback volume produces. Hierarchy correctness scores every individual edge in both trees as correct (Table~\ref{tab:existing}) because it never looks at how many siblings a node has.}
\label{fig:fanout}
\end{figure}

\textbf{Fan-out shape.} For each adjacent level pair, the distribution of $|\mathrm{children}(p)|$ over parents $p$. Two summary statistics are reported per artifact: \emph{max fan-out (any pair)}, the largest $|\mathrm{children}(p)|$ observed across every parent and every level pair; and \emph{cap-suspect level pairs}, the count of level pairs whose parent set $P$ satisfies
\[
|P| \geq 10 \quad\text{and}\quad \max_{p \in P} |\mathrm{children}(p)| \leq 3
\]
that is, at least ten parents and \emph{no} parent with more than three children -- a ceiling on the entire level, not a count of small parents (any long-tailed tree has many of those). Real feedback volume is long-tailed; a universal cap this narrow does not occur by chance. Figure~\ref{fig:fanout} shows both patterns from the actual corpus.

\textbf{Cross-branch reuse.} $|\{n : |\mathrm{parents}(n)| > 1\}|$, the count of nodes reachable through more than one parent. A taxonomy expressed as a strict tree with zero reuse anywhere is a signal that cross-cutting concerns were duplicated or never modeled, rather than shared. The small-sample reference, for example, shares the real node ``Pre-Execution Safety Controls For Risky Agent Actions'' under four L3 parents, one of its 48 shared nodes; in a zero-reuse tree the same concept must be authored once per branch, splitting its traffic across duplicates that no generic check can distinguish from an intentional, distinct pair.

\begin{table}[t]
\centering
\footnotesize
\setlength{\tabcolsep}{2pt}
\begin{tabular}{@{}lcccccc@{}}
\toprule
& \multicolumn{3}{c}{Small sample} & \multicolumn{3}{c}{Large sample} \\
\cmidrule(lr){2-4} \cmidrule(lr){5-7}
\textbf{Metric}\footnotemark & Live & C-1 & C-2 & Live & D-1 & D-2 \\
\midrule
Containment $\downarrow$ & 13.9\% & \textbf{97.7\%} & \textbf{100\%} & 2.9\% & \textbf{97.7\%} & \textbf{100\%} \\
Information gain $\uparrow$ & 0.76 & 0.33 & 0.34 & 0.81 & 0.74 & 0.58 \\
Type-token ratio $\uparrow$ & 0.25 & 0.10 & 0.06 & 0.14 & 0.07 & 0.05 \\
Cap-suspect pairs $\downarrow$ & 0 & 1 & 1 & 1 & 0 & 1 \\
Cross-branch reuse $\uparrow$ & 48 & 0 & 0 & 73 & 0 & 0 \\
Max fan-out $\star$ & 31 & 11 & 26 & 67 & 109 & 200 \\
\bottomrule
\end{tabular}
\caption{Structural discriminators at the leaf level, six artifacts across two corpora. Every generated artifact collapses on containment, information gain, diversity, and reuse; max fan-out ($\star$) is distributional context, not a quality score. The large-sample reference trips the cap-suspect flag once, at a genuinely narrow level pair (30 parents, none above 3 children) -- the threshold false positive anticipated in Limitations -- while D-2's flagged pair is degenerate (451 parents, none above 2). Bold marks the headline finding. \textbf{The generic checks flagged none of this.}}
\label{tab:structural}
\end{table}
\footnotetext{$\uparrow$ higher is better; $\downarrow$ lower is better; $\star$ no single best direction -- max fan-out reflects the shape of the underlying data, not quality directly (\S\ref{sec:newmetrics}).}

\textbf{Reading Table~\ref{tab:structural}.} Ancestor containment jumps from 13.9\% and 2.9\% to \textbf{97.7--100\%} as soon as we leave the references, independently on both corpora: every generated artifact builds leaf names by concatenation. Table~\ref{tab:existing}'s naming scores (0.97--1.00 for the generated artifacts) missed this entirely because concatenated names still describe their own node correctly; they just add nothing beyond what the ancestors already said. Type-token ratio collapses 2.5--4.4$\times$ on the small sample and 2.0--2.7$\times$ on the large. Notably, C-1 -- the generated artifact with a perfect 1.00 hierarchy-correctness score -- still shows a 2.5$\times$ diversity collapse, so edge-level correctness and label informativeness clearly measure different things. All four generated artifacts have zero cross-branch reuse, against 48 and 73 shared nodes in the references. None of these six rows exists among the generic checks, because none of them is a fact about one node.

\subsection{Team Partitionability}

A product-feedback taxonomy's top levels often double as an ownership map, with L1 (and L2) branches assigned to separate product teams. Whether any such assignment can work is, at bottom, a question of structural separability: do the branches partition the record stream into groups that stay apart? We answer it by routing each artifact's record classifications through its tree. All three metrics below therefore depend on the classification as well as the tree itself (see Limitations).

\textbf{Branch leakage.} Let $\mathrm{branch}(r)$ be the set of distinct branches record $r$'s predicted paths touch, and $\mathrm{count}(r,b)$ its number of paths landing in branch $b$:
\[
\mathrm{leakage} = \frac{|\{r : |\mathrm{branch}(r)| > 1\}|}{|\{r\}|}
\]
\[
\mathrm{purity} = \mathrm{mean}_r \; \frac{\max_b \mathrm{count}(r,b)}{\sum_b \mathrm{count}(r,b)}
\]
where \emph{record purity} is the mean share of a record's paths falling in its majority branch (cf.\ cluster purity in clustering evaluation \citep{Manning_Raghavan_Schutze_2008}). High leakage means a team owning one branch cannot see a full issue without involving a second team.

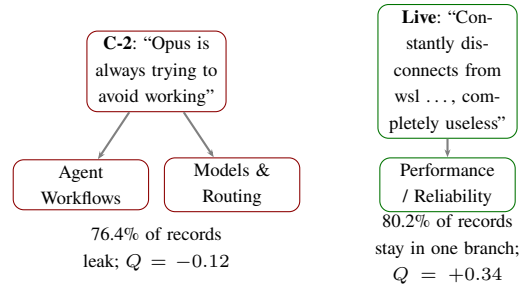
\begin{figure}[t]
\centering
\begin{tikzpicture}[
  every node/.style={font=\scriptsize},
  rbox/.style={draw, rounded corners, text width=1.7cm, align=center, inner sep=2pt},
  bbox/.style={draw, rounded corners, text width=1.65cm, align=center, inner sep=2pt},
  arr/.style={-{Stealth[length=3pt]}, thick, gray}
]
\node[rbox, draw=red!55!black] (r1) at (-2.3,0) {\textbf{C-2}: ``Opus is always trying to avoid working''};
\node[bbox, draw=red!55!black] (a1) at (-3.3,-1.5) {Agent Workflows};
\node[bbox, draw=red!55!black] (a2) at (-1.3,-1.5) {Models \& Routing};
\draw[arr] (r1)--(a1); \draw[arr] (r1)--(a2);
\node[font=\scriptsize, text width=2.3cm, align=center] at (-2.3,-2.35) {76.4\% of records leak; $Q=-0.12$};

\node[rbox, draw=green!45!black] (r2) at (1.55,0) {\textbf{Live}: ``Constantly disconnects from wsl \ldots, completely useless''};
\node[bbox, draw=green!45!black] (b1) at (1.55,-1.5) {Performance / Reliability};
\draw[arr] (r2)--(b1);
\node[font=\scriptsize, text width=2.3cm, align=center] at (1.55,-2.35) {80.2\% of records stay in one branch; $Q=+0.34$};
\end{tikzpicture}
\caption{Branch leakage and co-occurrence modularity ($Q$, Newman modularity of the branch partition: positive means branches hold together tighter than a random split, negative means worse). Both records and both quotes are real, taken verbatim from the corpus and its predictions; percentages give each pattern's share of that artifact's records. Left: a C-2 record about a specific model's behavior predicted onto both a model-routing branch and an agent-behavior branch, the majority pattern for C-2 (76.4\% of records leak), consistent with $Q<0$. Right: a Live record staying inside one branch, the majority pattern there (80.2\% of records touch exactly one branch), consistent with $Q>0$. Hierarchy correctness (Table~\ref{tab:existing}) scores C-1 and C-2 within 0.08 of each other; this figure's metrics separate them by \textbf{27 points of leakage and a sign flip}, because they are the only ones that route records through the tree at all.}
\label{fig:partition}
\end{figure}

\textbf{Co-occurrence modularity.} Build a weighted graph over leaf nodes where $w_{uv}$ is the number of records whose paths hit both leaves $u,v$; with $m = \sum_{uv} w_{uv}$ and branch degree $k_b = \sum_{u : c(u)=b} \sum_v w_{uv}$, Newman modularity \citep{Newman_2006} of the branch partition is
\[
Q = \sum_{b} \left[ \frac{e_b}{m} - \left(\frac{k_b}{2m}\right)^{2} \right]
\]
where $e_b$ is the intra-branch edge weight. $Q > 0$ means branches hold together tighter than a random partition; $Q < 0$ means the branch boundaries actively cut against how issues co-occur in practice -- worse than assigning branches at random. Figure~\ref{fig:partition} shows both failure modes on the same corpus.

\textbf{Dividability.} For branch $b$ with children volumes $\{v_1,\dots,v_k\}$ and total volume $V_b = \sum_i v_i$:
\[
H(b) = \frac{-\sum_i (v_i/V_b)\log(v_i/V_b)}{\log k}
\]
\[
\mathrm{div} = \frac{\sum_b V_b\, H(b)}{\sum_b V_b}
\]
$H(b) = 0$ for a single-child branch; $H(b) \to 1$ as volume splits evenly among children. This is a workload-balance diagnostic, not a quality signal by itself: an evenly split partition that is also fragmented (negative modularity) is not more useful, just more uniformly broken.

\begin{table}[t]
\centering
\footnotesize
\setlength{\tabcolsep}{2pt}
\begin{tabular}{@{}lcccccc@{}}
\toprule
& \multicolumn{3}{c}{Small sample} & \multicolumn{3}{c}{Large sample} \\
\cmidrule(lr){2-4} \cmidrule(lr){5-7}
\textbf{Metric}\footnotemark & Live & C-1 & C-2 & Live & D-1 & D-2 \\
\midrule
L1 leakage $\downarrow$ & 19.8\% & \textbf{49.8\%} & \textbf{76.4\%} & 19.3\% & 34.7\% & 39.3\% \\
L1 record purity $\uparrow$ & 0.92 & 0.78 & 0.61 & 0.93 & 0.83 & 0.81 \\
L1 modularity $\uparrow$ & 0.34 & \textbf{0.12} & \textbf{$-$0.12} & 0.43 & 0.08 & 0.16 \\
L1 dividability $\star$ & 0.32 & 0.87 & 0.71 & 0.70 & 0.59 & 0.62 \\
L2 leakage $\downarrow$ & 22.0\% & 68.1\% & 79.4\% & 22.6\% & 38.1\% & 43.7\% \\
L2 record purity $\uparrow$ & 0.91 & 0.58 & 0.60 & 0.91 & 0.79 & 0.77 \\
L2 modularity $\uparrow$ & 0.31 & 0.01 & $-$0.04 & 0.44 & 0.05 & 0.13 \\
L2 dividability $\star$ & 0.56 & 0.74 & 0.00 & 0.61 & 0.72 & 0.75 \\
\bottomrule
\end{tabular}
\caption{Team partitionability, six artifacts, each scored over its own classification of its corpus's records. On both corpora the references leak about a fifth of records across top-level branches; \textbf{every generated artifact leaks between a third and three-quarters}. Bold marks the C-1 vs.\ C-2 contrast discussed in the text. Dividability ($\star$) is workload-balance context, not a quality score.}
\label{tab:partition}
\end{table}
\footnotetext{$\uparrow$ higher is better; $\downarrow$ lower is better; $\star$ no single best direction -- see the dividability definition.}

\textbf{Reading Table~\ref{tab:partition}.} The references' L1 leakage (19.8\% and 19.3\%, Table~\ref{tab:partition}, row 1) means four in five records resolve inside a single product area on both corpora: a team owning that area sees whole issues, not fragments. Every generated artifact leaks 1.8--3.9$\times$ more (34.7--76.4\% against 19.3--19.8\%, same row); C-2's 76.4\% means three in four records span more than one team's branch. Modularity sits 2.6--5.6$\times$ below the references everywhere it stays positive, and turns \emph{negative} for C-2 at both levels: its branch boundaries are worse than a random split at keeping co-occurring concerns together, meaning the L1 labels look like categories on paper but do not correspond to how issues actually cluster in the feedback stream. One mechanism is visible in the labels themselves: C-1's L1 ``Outcomes, Adoption and Market Perception'' bundles unrelated concerns -- community discussion, competitor comparisons, industry direction, overall product experience -- that no single team could own, where the reference's L1s (e.g.\ Performance / Reliability, Cloud Execution) are each one coherent area. Hierarchy correctness cannot flag this, because every child edge is individually defensible; the category itself is a combination of topics standing in for a product area. Dividability illustrates why it is reported as context: balance without containment is not separability, and C-2 combines the study's worst leakage with an L2 dividability of exactly zero (every one of its 158 L2 parents has a single child, so there is nothing to divide). This is the sharpest disagreement with Table~\ref{tab:existing}. Hierarchy correctness scores C-1 (1.00) within 0.08 of C-2 (0.92), but partitionability separates them by \textbf{27 points of leakage and a modularity sign flip}: the edges are individually well formed in both trees, yet the branches behave very differently. An evaluation limited to generic checks cannot recover this because it never constructs the object -- the leaf co-occurrence graph -- that the comparison depends on.

\section{Discussion}

Every failure class in Sections~\ref{sec:existing}--\ref{sec:newmetrics} is a fact about the \emph{ensemble} of nodes or records, not about any node judged in isolation. Ancestor-containment collapse lives in the relationship between every leaf and its ancestors; branching caps in the distribution of children across dozens of parents; leakage and modularity in how thousands of record-to-path assignments co-occur across the corpus. An LLM judge invoked per node, per edge, or per sibling pair is architecturally unable to compute any of these -- not because the judge is weak, but because the question is not local. Our two families are cheap for the same reason: they need no judge, only closed-form statistics over the artifact and its classifications, and every number in Tables~\ref{tab:structural} and~\ref{tab:partition} was produced without an LLM call.

The metrics also settle the mutual-exclusiveness question deferred from Section 2.4. Mutual exclusiveness \citep{Nickerson2013AMF} is a node-level design prescription; branch leakage operationalizes it at the record level, where a record spanning two branches is a countable violation. Reuse and leakage are therefore complementary rather than contradictory: reuse asks whether a cross-cutting concept is represented once and shared, leakage asks how much of the record stream such concepts affect. A well-shaped taxonomy has nonzero reuse and low leakage (the references: 48 and 73 shared nodes, about 20\% leakage). The generated artifacts combine zero reuse with 35--76\% leakage -- cross-cutting concepts were neither shared nor contained.

The four runs never saw the evaluator, so nothing was optimized against it: templated labels, capped branching, and per-concept branch minting are what unconstrained generation produces by default, and generic scoring fails to detect them. A process optimized \emph{against} such checks would, per Goodhart's Law, drift toward them. More compute does not help either -- the heaviest run on each corpus (C-2 at 9.5M input tokens, D-2 at 12.2M) is also its corpus's worst on leakage and diversity. Coverage checks turn the missing penalty into an active reward, since minting a node per documented term is the cheapest way to raise coverage and manufactures these same patterns; the tree with the study's worst leakage (C-2) posts the highest small-sample product-term coverage (0.86 vs.\ 0.72), and the pattern repeats on the large sample (0.90--0.91 vs.\ 0.80). A team relying on generic checks to gate taxonomy quality should add a corpus-level pass, whether the two families here or equivalents, before trusting the result.

\section{Limitations and Future Work}

All generated runs use one harness, so harness diversity is untested, though the mechanism (corpus-level facts invisible to generic judging) is not harness-specific. The cap-suspect threshold can false-positive on genuinely narrow branching, and does once, on the large-sample reference (Table~\ref{tab:structural}). Type-token ratio is size-sensitive across very different scales. Leakage, purity, and modularity depend on each artifact's own record classification, so they evaluate the taxonomy-plus-classifier system. Neither family is validated against an external usefulness signal such as team velocity; we claim they detect failures existing metrics miss here, not that they suffice alone. Every metric in Table~\ref{tab:existing} is LLM-judged (gpt-5.4-mini; gpt-5.5 for coverage) without a human-agreement study \citep{10.1162/coli.07-034-R2}, and such judges show self-preference \citep{Zheng2023JudgingLW,Panickssery2024LLMER} and order sensitivity \citep{shi2025judgingjudgessystematicstudy}; our judge-free families are unaffected. Future work includes temporal stability, a second harness family, tuned thresholds, and insertion-time construction gates.

\paragraph{Availability and ethics.} The corpora, artifacts, and documentation are proprietary customer data and are not released, so Tables~\ref{tab:structural} and~\ref{tab:partition} cannot be independently recomputed. Every metric is fully specified in Section~\ref{sec:newmetrics}'s formulas and Section~\ref{sec:setup}'s protocol, and the complete agent task specification and launch prompt are given in the Supplementary Material, so both metric families and the construction protocol can be reimplemented from the paper alone and applied unchanged to any taxonomy artifact meeting the same input contract.

\section{Conclusion}

A taxonomy can clear every LLM-judged generic check in common use, and beat its production reference on product-term coverage, while every leaf label restates its ancestors verbatim, its branching is generator-capped, and three-quarters of its records leak across the product-area boundaries it exists to enforce. We observed this in all four runs of the harness we tested, independently on two real corpora: the same failure modes, every run. Structural discriminators and team partitionability recover the failures generic judging cannot see, deterministically and with no LLM judge. The metric a team is missing is not a better generic judge; it is a metric that looks at the tree, and the records routed through it, as a whole.

\bibliography{references}

\end{document}